\documentclass[10pt,twocolumn,letterpaper]{article}
\usepackage{cvpr}
\usepackage{times}
\usepackage{epsfig}
\usepackage{graphicx}
\usepackage{amsmath}
\usepackage{amssymb}
\usepackage{booktabs}
\usepackage{amssymb}
\usepackage{pifont}
\usepackage[pagebackref=true,breaklinks=true,letterpaper=true,colorlinks,bookmarks=false]{hyperref}

\cvprfinalcopy

\def\cvprPaperID{19} 

\ifcvprfinal\fi
\begin{document}

\title{Dropout as an Implicit Gating Mechanism For Continual Learning}

\author{Seyed Iman Mirzadeh\\
Washington State University\\
{\tt\small seyediman.mirzadeh@wsu.edu}\\
\and
Mehrdad Farajtabar\\
DeepMind\\
{\tt\small farajtabar@google.com}\\
\and
Hassan Ghasemzadeh\\
Washington State University\\
{\tt\small hassan.ghasemzadeh@wsu.edu}\\
}

\maketitle

\begin{abstract}
  In recent years, neural networks have demonstrated an outstanding ability to achieve complex learning tasks across various domains. However, they suffer from the ``catastrophic forgetting'' problem when they face a sequence of learning tasks, where they forget the old ones as they learn new tasks. This problem is also highly related to the ``stability-plasticity dilemma''. The more plastic the network, the easier it can learn new tasks, but the faster it also forgets previous ones.
  Conversely, a stable network cannot learn new tasks as fast as a very plastic network. However, it is more reliable to preserve the knowledge it has learned from the previous tasks. 
  Several solutions have been proposed to overcome the forgetting problem by making the neural network parameters more stable, and some of them have mentioned the significance of dropout in continual learning. However, their relationship has not been sufficiently studied yet.
  In this paper, we investigate this relationship and show that a stable network with dropout learns a gating mechanism such that for different tasks, different paths of the network are active. 
  Our experiments show that the stability achieved by this implicit gating plays a very critical role in leading to performance comparable to or better than other involved continual learning algorithms to overcome catastrophic forgetting.\footnote{The code and the appendix is available at: \tt \scriptsize https://github.com/imirzadeh/stable-continual-learning}
\end{abstract}


\section{Introduction}

\begin{figure}[t!]
    \centering
    \includegraphics[width=0.85\linewidth]{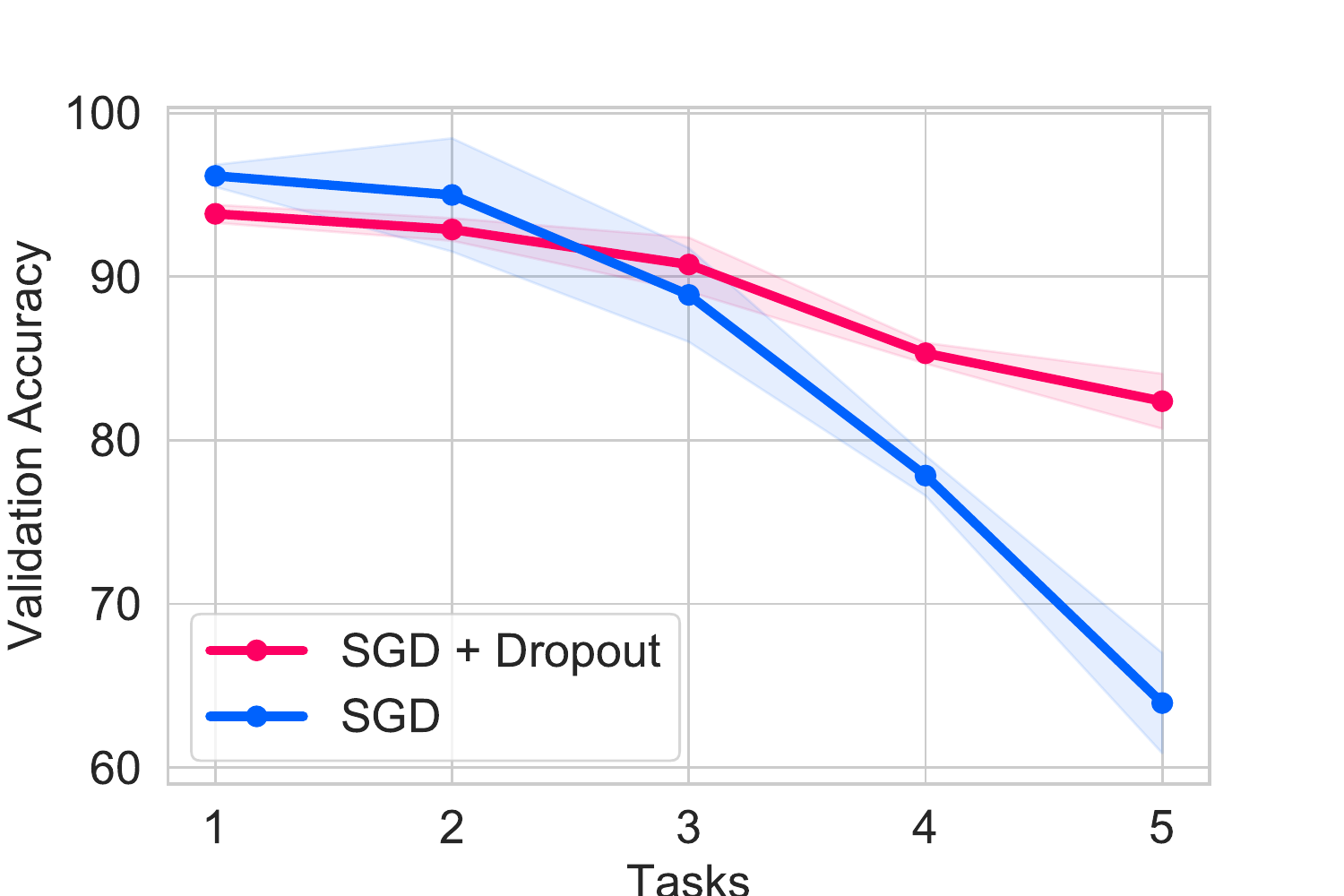}
    \caption{Networks trained with dropout tend to forget at a slower rate. The lines represent the evolution of the validation accuracy of the first task, as networks learn new tasks}
    \label{fig:intro-problem}
\end{figure}

The stability-plasticity dilemma is a well-known problem for both artificial and biological neural networks~\cite{stability-plasticity}. Intelligent systems need \textit{plasticity} to learn new knowledge and adapt to new environments while they require \textit{stability}  to prevent forgetting previous knowledge.
If a network is very plastic but not stable, it can learn new tasks faster, but it also forgets the previous ones easily. This is known as the catastrophic forgetting problem ~\cite{McCloskey1989CatastrophicII}. 
On the other hand, a network can be very stable and preserves the knowledge of the previous tasks, but it cannot easily adapt to unseen environments and learn new tasks.

We motivate our paper by illustrating the stability-plasticity dilemma in a standard continual learning dataset in Figure~\ref{fig:intro-problem}. The tasks in this dataset are generated by continually rotating the MNIST digits. The red and blue lines represent the two algorithms, respectively: (1) Stochastic Gradient Descent (SGD) with Dropout~\cite{DROPOUT_ORIGINAL} and (2) SGD without Dropout.
The network trained without dropout can quickly pick up new tasks (plasticity); however, forgets previous ones as we move forward to subsequent tasks. On the other hand, the network that is trained with dropout retains the previous knowledge significantly better (stability) by paying a small cost of performance drop.

To the best of our knowledge, the work by~\cite{Goodfellow2013AnEI} is the first to empirically study the importance of the dropout technique in the continual learning setting. 
They hypothesize that dropout increases the optimal size of the network by regularizing and constraining the capacity to be just barely sufficient to
perform the first task. However, by observing some inconsistent results on dissimilar tasks, they suggested dropout may have other beneficial effects too.
More recently, the effectiveness of dropout is demonstrated in a comprehensive study on several architectures and datasets~\cite{Lange2019ContinualLA,Zenke2017ContinualLT}.
However, many important questions about the relationship between the dropout method and catastrophic forgetting are unanswered. One such question is ``\textit{How does the dropout help the network to overcome the catastrophic forgetting besides regularization?}''. It is well established that dropout works as a regularizer~\cite{ExpImpDropout}. But, several other regularizers (\textit{e.g.}, L2 norm) fail to help the network in a continual learning setting\cite{EWC}.

In this paper, we analyze the impact of dropout on network stability and study its behavior in the presence of dropout. 
We show that the dropout networks behave like a network with a gating mechanism, and the crated task-specific pathways are retained and consistent during the sequential learning of tasks. 
Finally, we show that training with dropout gives a stable and flexible network that outperforms several other methods when they do not use dropout even if they are equipped with an external memory of previous examples.

\section{Related Work}
Several continual learning methods have been proposed to tackle catastrophic forgetting. We can categorize these algorithms into three general groups, followed by by~\cite{Lange2019ContinualLA}. 

The first group consists of replay based methods that build and store a memory of the knowledge learned from old tasks, known as \textit{experience replay}. iCaRL~\cite{Rebuffi2016iCaRLIC} learns in a class-incremental way by having a fixed memory that stores samples that are close to the center of each class. Averaged Gradient Episodic Memory (A-GEM)~\cite{AGEM} is another example of these methods which build a dynamic episodic memory of parameter gradients during the learning process. Very recently, the Hindsight Anchor Learning (HAL)~\cite{Chaudhry2019HAL} proposed to keep some ``anchor'' points of past tasks and use these points to update knowledge on the current task.

The methods in the second group use explicit regularization techniques to supervise the learning algorithm such that the network parameters are consistent during the learning process. Elastic weight consolidation (EWC)~\cite{EWC} uses the Fisher information matrix as a proxy for weights' importance and guide the gradient updates. Orthogonal Gradient Descent (OGD)~\cite{OGD} uses the projection of the prediction gradients from new tasks on the subspace of previous tasks' gradients to protect the learned knowledge. The idea of using knowledge distillation~\cite{Hinton2015KD,Mirzadeh2019ImprovedKD,Mobahi2020SelfDistillationAR} is also found to be a successful regularizer in several works~\cite{Li2018LWF,lee2019overcomingKDCL}.

Finally, in parameter isolation methods, in addition to potentially a shared part, different subsets of the model parameters are dedicated to each task. This approach can be viewed as a flexible \textbf{gating mechanism}, which enhances stability and controls the plasticity by activating different gates for each task.
\cite{Gating} proposes a neuroscience-inspired method for a context-dependent gating signal, such that only sparse, mostly non-overlapping patterns of units are active for any one task. PackNet~\cite{PackNet} implements a controlled version of gating by using network pruning techniques to free up parameters after finishing each task and thus sequentially “pack” multiple tasks into a single network.
Gating mechanisms found to be very efficient in several works. In the comprehensive set of experiments, PackNet is shown one of the most reliable methods~\cite{Lange2019ContinualLA} and adding the context-dependent-gates to other methods such as EWC improved their performance drastically~\cite{Gating}.

In the following sections, we show that a stable network trained with dropout will learn a reliable gates mechanism. We note that the majority of the mentioned methods need extra computation and memory costs, while a stable dropout network is a much more memory and computation efficient.

\section{Dropout and Network Stability}
\label{sec:dropout}
Dropout~\cite{DROPOUT_ORIGINAL} is a well-established technique in deep learning, which is also well-studied theoretically ~\cite{DROPOUT_UNDERSTANDING,DROPOUT_THESIS,Helmbold2016SurprisingPO,ExpImpDropout}. It was originally designed to prevent the co-adaptation of neurons in a network. It increases the stability of neural networks and has been employed successfully in various domains. In the training phase of a dropout network, at each example presentation, neurons are deleted with probability $1-p$, and the network will be trained in a standard way. For the inference phase, the weights are re-scaled proportional to the dropout probability.

There are various viewpoints to dropout. In this paper, we are interested in regarding dropout as a method for sparse coding and regularization and leveraging the associated theoretical insights to study the relationship between dropout and continual learning.

Consider the neuron $i$ of the layer $h$ in a neural network and define the activity of the neuron by: \\
\begin{align}
    S_i^{h} = \sum_{l<h}{\sum_{j}{w_{ij}^{hl} S_j^l} \delta_{j}^{l}} \quad S_j^0 = I_j,
    \label{eq:dropout-activation}
\end{align}
where, $I$ is the input vector and $w_{ij}^{hl}$ represents weight from neuron $j$ of layer $l$ to neuron $i$ of layer $h$. $\delta_j^l$ is the \textbf{gating} binary Bernoulli random variable which is indicating if the neuron is disabled by the dropout (i.e., $P(\delta_j^l = 1) = p_j^l$) or not. Under the assumption that $\delta_j^l$'s are independent, and dropout has not been applied to previous layers, \cite{DROPOUT_UNDERSTANDING} showed that if we apply dropout to layer $h$ the variance of the activation for each neuron follows: 
\begin{align}
    \text{Var}(S_i^h) = \sum_{l < h} (w_{ij}^{hl})^2 \sigma(S_j^l)^2 p_j^l (p_j^l-1).
    \label{eq:dropout-sparse-coding}
\end{align}
Where $\sigma(S_j^l)$ denotes the output of neuron $j$ at previous layer $l$. Therefore, to obtain a stable activation behavior, the variance of the activation of a neuron should be minimized. This happens if $p_j^l$ is close to either $0$ or $1$ (so $ p_j^l (p_j^l-1)$ will be small). Note that we can not directly control $w$ as it will be updated by the loss function objective.

One consequence of Equation~\eqref{eq:dropout-sparse-coding} is that in a stable dropout network, the neural activation is very sparse. This yields to a skewed asymmetric distribution for neuron activity inside a network~\cite{DROPOUT_UNDERSTANDING}. 
This skewed asymmetric distribution has close connections to the gating mechanism. Such a distribution for the neural activity of several animals is believed to be responsible for an optimal trade-off between stability and plasticity~\cite{BrainLogDynamic}. This firing pattern implements a gating mechanism inside the brain that is not only plastic enough to learn new tasks but also stable enough to preserve the knowledge it has learned from different tasks. \cite{BrainLogDynamic} showed the neural activity of several biological brains, which is in line with the neural activity in dropout networks, as shown by ~\cite{DROPOUT_UNDERSTANDING} and our experiments. 

Training with dropout also has another consequence: dropout most heavily regularizes the neurons that contribute to uncertain predictions (i.e., semi-active neurons that are not close to either 0 or 1)~\cite{ExpImpDropout,DROPOUT_UNDERSTANDING}. Intuitively, for a network of gates and switches, it means that dropout regularization pushes neurons to be either active or deactivate. EWC~\cite{EWC} also is built upon the same intuition of penalizing the changes to certain weights and allowing the less certain parameters to handle learning new tasks. 
When the model finishes task $t$ and reaches task $t+1$, this regularization would create new gates by either enabling or disabling such neurons. Decaying the learning rate during the continual learning experience also helps dropout increase the model stability since by preserving the gates for a longer time.  

In conclusion, dropout regularization helps to create gates in the network by pushing the neurons to be either highly active or highly inactive during the learning experience. In addition, when facing new tasks, the regularization mechanism will change the semi-active neurons more compared to active or inactive neurons, which helps to preserve the task-specific pathways when learning subsequent tasks.

\section{Experimental Setup}
\subsection{Datasets}
We perform our experiments on two standard continual learning benchmarks: Permuted MNIST~\cite{Goodfellow2013AnEI}, and Rotated
MNIST.
Each task of the permuted MNIST dataset is generated by shuffling the pixels of images such that the permutation would be the same between images of the same task but is different across the tasks. 
Each permutation is chosen randomly; thus, the difficulty of tasks is the same. We used the first task to be the original MNIST images.
Rotated MNIST is generated by the continual rotating of the original MNIST images. Here, task 1 is to classify standard MNIST digits, and each subsequent task will rotate the previous task's images by 10 degrees (e.g., task 2 by 10 degrees, task 3 by 20 degrees, and so on).

\subsection{Training Settings}
In this section, we first describe our training setting for Sections~\ref{sec:experiments-plasticity-stability}, \ref{sec:experiments-dropout-gating}, and \ref{sec:experiments-compare}.  
    We use PyTorch~\cite{Pytorch} for the implementation of all experiments and reported the average and standard deviation of the validation accuracy for five runs. For all experiments, we use a multi-layer perceptron (MLP) with two hidden layers, each with 100 ReLU neurons. Moreover, each network is only trained on each task for five epochs to be consistent with several other benchmarks~\cite{OGD}. We compare the standard SGD training with Elastic Weight Consolidation (EWC)~\cite{EWC}, A-GEM~\cite{AGEM}, and Orthogonal Gradient Descent (OGD)~\cite{OGD}. Multi-Task Learning (MTL) serves as an upper bound and that the network is trained in a multi-task setting (i.e., data of previous tasks are always available and used for training). All the results except the SGD with dropout were directly cited from~\cite{OGD} as datasets, training epochs, and optimizers were the same. For SGD with dropout, we use the batch size of 64 and the standard SGD optimizer with a learning rate of $0.01$ and $0.8$ for momentum. Furthermore, we found that learning rate decay helps network stability dramatically, and we reduced the learning rate by $0.8$ after finishing each task. We have experimented with different dropout probabilities and found that values between $0.2$ and $0.6$ work well. However, for simplicity, we have used $0.5$ for the dropout probability for all reported results unless stated otherwise.  We note that all the methods except the SGD+Dropout are trained without dropout and learning rate decay since our main goal is to measure the performance gain of the methods that are not due to these stability techniques.
\begin{figure*}[t!]
    \centering
    \includegraphics[width=0.9\textwidth]{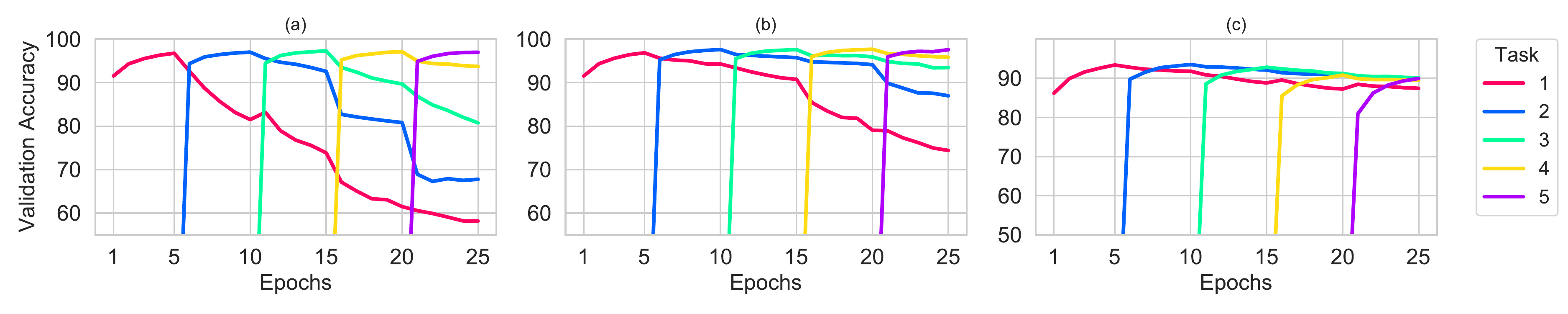}
    \caption{\textit{Permuted MNIST}- Increasing the stability and reducing the plasticity from left to right by increasing the the dropout rate and learning rate decay.}
    \label{fig:compare-stablity-permuted}
\end{figure*}
\begin{figure*}[t!]
    \centering
    \includegraphics[width=0.9\textwidth]{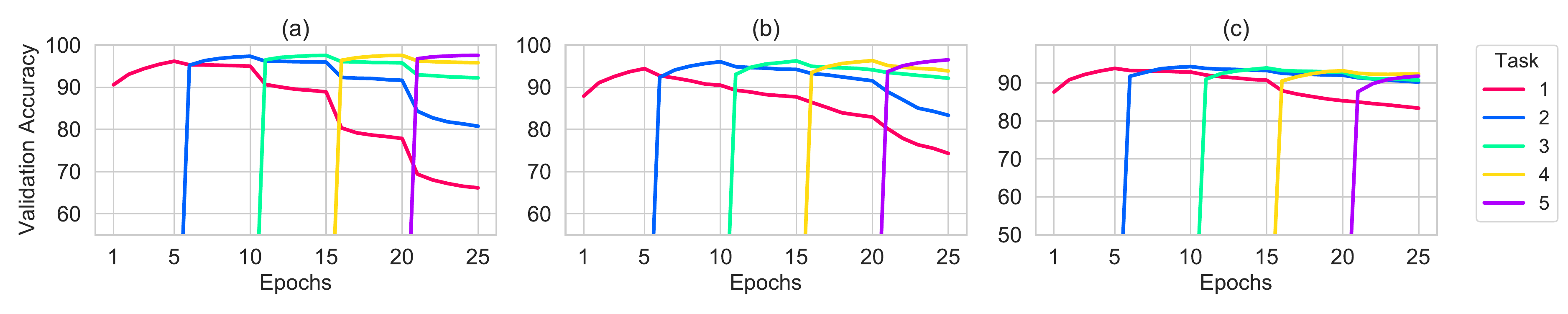}
    \caption{\textit{Rotated MNIST}- Increasing the stability and reducing the plasticity from left to right by increasing the the dropout rate and learning rate decay.}
    \label{fig:compare-stablity-rotated}
\end{figure*}

For our scaled experiment (Section ~\ref{sec:experiments-scaled}), we extend the number of tasks to 20 rather 5  to verify our analysis holds. We used a two-layer MLP with 256 ReLU neurons in each layer. For each task, the network will be trained for 5 epochs. The dropout parameter and learning rate decay will remain the same as the previous section. For this experiment, we use two metrics from~\cite{Chaudhry2018RiemannianWF,AGEM} to evaluate continual learning algorithms when the number of tasks is large:
\begin{enumerate}
    \item \textbf{Average Accuracy}: The average validation accuracy after the model has been trained sequentially up to task $t$, defined by:
    \begin{align}
        A_t = \frac{1}{t} \sum_{i=1}^t a_{t,i},
        \label{eq:average-acc}
    \end{align}
    where, $a_{t,i}$ is the validation accuracy on dataset $i$ when the model finished learning task $t$.
    
    \item \textbf{Forgetting Measure}: The average forgetting after the model has been trained sequentially on all tasks. Forgetting is defined as the decrease in performance at each of the tasks between their peak accuracy and their accuracy after the continual learning experience has finished. For a continual learning dataset with $T$ sequential tasks, it is defined as:
    \begin{align}
        F = \frac{1}{T-1} \sum_{i=1}^{T-1}{\text{max}_{t \in \{1,\dots, T-1\}}~(a_{t,i}-a_{T,i})}.
        \label{eq:forgetting-measure}
    \end{align}
\end{enumerate}

Finally, in our code repository, we provide scripts to reproduce the results with suggested hyper-parameters.

\section{Results}
In this section, we perform several experiments to show the impact of dropout on model stability.

\subsection{Forgetting Curve in Stable Networks}
\label{sec:experiments-plasticity-stability}
In our first experiment, we show that it is feasible to increase the stability of a network by compromising its plasticity a little bit but getting a considerable amount of stability in return.
Figures~\ref{fig:compare-stablity-permuted} and~\ref{fig:compare-stablity-rotated} show the evolution of validation accuracy throughout the continual learning over five tasks on permuted MNIST and rotated MNIST, respectively. For each dataset, we train networks for three different settings:  
\begin{itemize}
    \item \textbf{(a)} Training the network without dropout and learning rate decay and obtain a \textit{highly plastic} network.
    \item \textbf{(b)} Training the network with small dropout probability ($p=0.25$) and also learning rate decay to obtain a more stable network than the one in part(a). 
    \item \textbf{(c)} Training with moderate dropout ($p=0.5$) and applying learning rate decay which yields a \textit{highly stable} network.
\end{itemize}
We would like to clarify that the x-axis in both figures denotes the time in the continual learning experience. Since the learning experience consists of five tasks, each for five epochs, the x-axis time denotes the time, which would be between one and twenty-five. The reported numbers at each step are calculated by averaging the accuracy over five different runs.
We can observe from Figures~\ref{fig:compare-stablity-permuted} and~\ref{fig:compare-stablity-rotated} that plastic networks in (a) learn new tasks better and faster than more stable ones, but they also forget old tasks very quickly. Networks with moderate plasticity in (b) learn slower than the highly plastic ones in (a), but they also forget at a slower rate. Finally, highly stable networks in (c) have the slowest forgetting curve thanks to the switching gates of the dropout. However, the stability comes with its cost: compromising flexibility, which yields to learning new tasks at a slower rate.

We emphasize our main goal of this experiment: It is possible to obtain stable networks by compromising the right amount of plasticity, and unlike OGD~\cite{OGD} and AGEM~\cite{AGEM}, with no additional techniques such as replay memory and correcting gradient directions. We will see in Section~\ref{sec:experiments-compare} that improving stability plays a much more important role than the mentioned techniques. We will show that stable networks trained with SGD can outperform other continual learning methods when they do not exploit these simple yet effective stability techniques.

\subsection{Dropout and Gating Mechanism}
\label{sec:experiments-dropout-gating}

\begin{figure*}[t!]
    \centering
    \includegraphics[width=\textwidth]{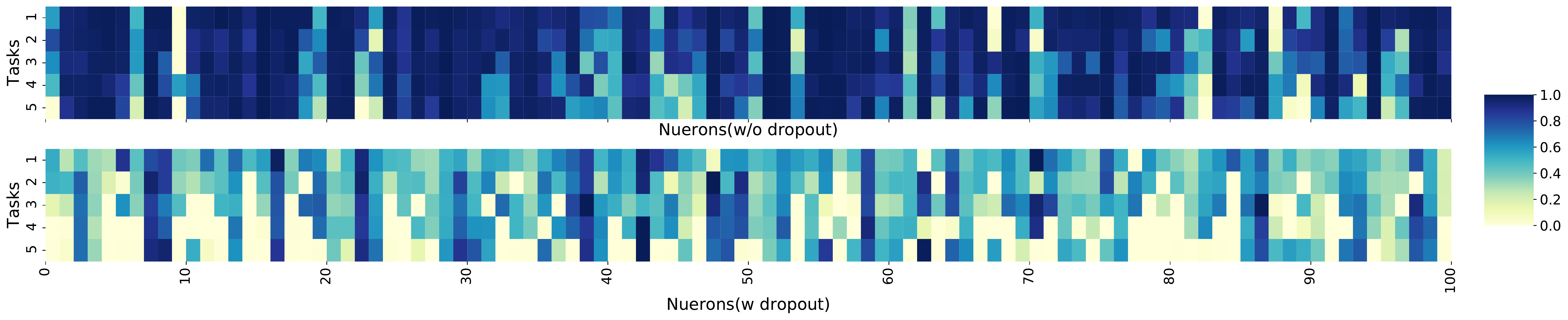}
    \caption{The effect of dropout on the activation(firing) pattern of neurons}
    \label{fig:firing-patterns}
\end{figure*}

\begin{figure*}[t!]
    \centering
    \includegraphics[width=\textwidth]{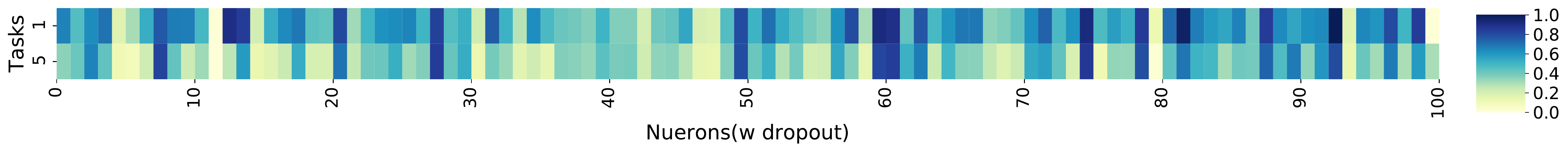}
    \caption{\textit{Stable Network} - Consistency between activation patterns of neurons for task 1, after learning task 1 and task 5 }
    \label{fig:consistent-firing}
\end{figure*}

In this experiment, we show that training with dropout will implicitly produce different gates in the network such that for each task, only a certain subset of the network parameters is active. 

 We counted the number of times a specific neuron was active (fired) for each task, and compare this behavior throughout the sequential learning process for all the tasks. Figure~\ref{fig:firing-patterns} shows the heatmap of the activation behavior of neurons of the first layer of two networks (with and without dropout) that are trained using the SGD method after finishing five permuted MNIST tasks. 
We indexed 100 neurons on the x-axis (from 0 to 99) and plotted the heatmap of their activation on the y-axis indexed by tasks. In other words, it represents the frequency of activation on validation data of that task. We note that the number of times that a neuron can fire for each MNIST task will be between  0 and 10000 (size of validation set). For better representation, we have normalized this number by dividing each value by 10000 so that all numbers are between 0 and 1. 

The first interesting observation from Figure~\ref{fig:firing-patterns} is that the activity pattern of neurons of the network trained with dropout is sparser than the case without dropout. Some neurons are very active, and some very inactive. 
This is in contrast to the behavior of the network without dropout, where almost all the neurons are very active for all tasks.
Moreover, if we focus on the behavior of a single neuron of a network with dropout, we see that the neuron is active for some tasks but is inactive for the others. Only a few of them are always active for all tasks. 
This behavior shows the gating mechanism of the network trained with dropout. 
The second interesting observation is the evolution of activation sparsity as the model learns more and more tasks. In other words, fewer and fewer neurons remain free to be activated for later tasks. This is due to the fact that the network's remaining capacity fills up as training continues.

It's notable that the gating mechanism is most useful when the pathways for a task remains consistent and almost invariant while training on subsequent tasks and so on. 
When the network is learning task $t$, dropout helps to produce some gating for the forward propagation. However, if the gates for this task are not preserved throughout the sequential learning process and change while the network is learning task $t+1$, then the network will forget task $t$. 
Figure~\ref{fig:consistent-firing} shows the activation patterns of task 1 for the first layer of a network trained with dropout, at two different times: (1) right after learning the first task (beginning of the continual learning), and (2) after learning the final task (end of the learning). As illustrated, the activation behavior and the gating is fairly consistent, and the pathways are preserved through time.

Finally, we note that although the illustrated examples are only for five tasks of permuted MNIST, the same pattern of behavior exists for the networks trained on the rotated MNIST task and when the number of tasks increases.

\subsection{Increasing Tasks}
\label{sec:experiments-scaled}
\begin{figure}[t!]
    \centering
    \includegraphics[width=0.8\linewidth]{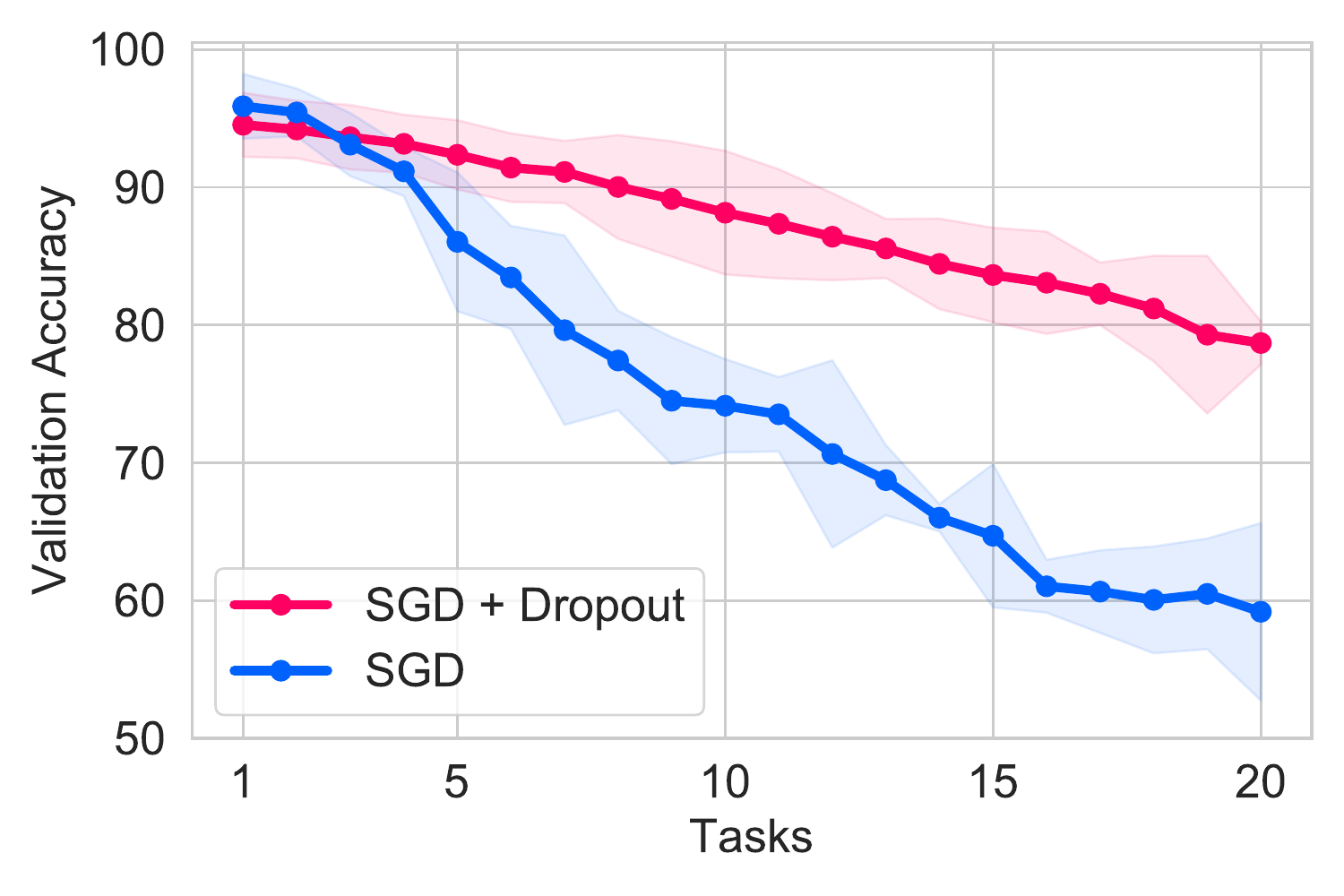}
    \caption{Comparison of average accuracy at the end of each task for several methods}
    \label{fig:compare-scaled-permuted}
\end{figure}

In this section, we show that the stability of dropout training remains effective in the case of an increased number of tasks.

Figure~\ref{fig:compare-scaled-permuted} compares the evolution of average accuracy (Equation ~\eqref{eq:average-acc}) for a stable versus plastic network. The graph consists of the average and three standard deviations over five different runs.
The stable networks have the final average accuracy of $78.7\;(\pm 0.2)$ with forgetting statistic (Equation~\eqref{eq:forgetting-measure}) of $0.13\;(\pm 0.02)$ while these metrics for plastic networks are $59.2\;(\pm 2.7)$  and $0.39\;(\pm 0.03)$, respectively.

In the appendix section, we compare the stable dropout networks with various state of the art continual learning settings for 20 tasks.

\subsection{Comparison with Other Methods}
\label{sec:experiments-compare}
In this experiment, we compare the stable SGD+Dropout network with several other continual learning methods. The goal is to compare the significance of the \textit{``network stability''} compared to the methods that focus on the other aspects of the catastrophic forgetting to tackle this problem.

Table~\ref{tab:permutation-mnist} and~\ref{tab:rotated-mnist} compare several continual learning algorithms on the permuted and rotated MNIST datasets. SGD+Dropout outperforms all the continual learning methods on old tasks and achieves an acceptable accuracy on new tasks for both permuted and rotated MNIST datasets. One interesting observation from both tables is the fact that SGD+Dropout achieves near-optimal accuracy even on new tasks but is not the best. The reason is that the network is very stable, and because of the stability-flexibility trade-off, it has lost some part of its flexibility.

Finally, we note that all the other methods except SGD, EWC, and SGD+Dropout are using some 200 data points per task to calculate gradient information from previous tasks (e.g., OGD) or in the form of episodic memory (e.g., A-GEM).

\begin{table}[t!]
\centering
\resizebox{\linewidth}{!}{%
\begin{tabular}{l|ccccc}
\hline
 & \multicolumn{5}{c|}{\textbf{Accuracy $\pm$ std (\%)}} \\ \hline
 & \textbf{Task 1} & \textbf{Task 2} & \textbf{Task 3} & \textbf{Task 4} & \textbf{Task 5} \\ \hline
MTL & 93.2 $\pm$ 1.3 & 91.5 $\pm$ 0.5 & 91.3 $\pm$ 0.7 & 91.3 $\pm$ 0.6 & 88.4 $\pm$ 0.8 \\ \hline
OGD & 79.5 $\pm$ 2.3 & 88.9 $\pm$ 0.7 & 89.6 $\pm$ 0.3 & \textbf{91.8 $\pm$ 0.9} & 92.4 $\pm$ 1.1 \\
A-GEM & 85.5 $\pm$ 1.7 & 87.0 $\pm$ 1.5 & 89.6 $\pm$ 1.1 & 91.2 $\pm$ 0.8 & \textbf{93.9 $\pm$ 1.0} \\
EWC & 64.5 $\pm$ 2.9 & 77.1 $\pm$ 2.3 & 80.4 $\pm$ 2.1 & 87.9 $\pm$ 1.3 & 93.0 $\pm$ 0.5 \\
SGD & 60.6 $\pm$ 4.3 & 77.6 $\pm$ 1.4 & 79.9 $\pm$ 2.1 & 87.7 $\pm$ 2.9 & 92.4 $\pm$ 1.1 \\
SGD+Dropout & \textbf{88.2 $\pm$ 1.6} & \textbf{90.3 $\pm$ 1.1} & \textbf{91.2 $\pm$ 2.0} & 90.3 $\pm$ 1.2 & 89.9 $\pm$ 1.4 \\ \hline
\end{tabular}%
}
\caption{\textit{Permuted MNIST}: The validation accuraficy of the model for each task, after being trained on all tasks in sequence.}
\label{tab:permutation-mnist}
\end{table}

\begin{table}[t!]
\centering
\resizebox{\linewidth}{!}{%
\begin{tabular}{l|ccccc}
\hline
 & \multicolumn{5}{c|}{\textbf{Accuracy $\pm$ std (\%)}} \\ \hline
 & \textbf{Task 1} & \textbf{Task 2} & \textbf{Task 3} & \textbf{Task 4} & \textbf{Task 5} \\ \hline
MTL & 92.1 $\pm$ 0.9 & 94.3 $\pm$ 0.9 & 95.2 $\pm$ 0.9 & 93.4 $\pm$ 1.1 & 90.5 $\pm$ 1.5 \\ \hline
OGD & 75.6 $\pm$ 2.1 & 86.6 $\pm$ 1.3 & 91.7 $\pm$ 1.1 & 94.3 $\pm$ 0.8 & 93.4 $\pm$ 1.1 \\
A-GEM & 72.6 $\pm$ 1.8 & 84.4 $\pm$ 1.6 & 91.0 $\pm$ 1.1 & 93.9 $\pm$ 0.6 & \textbf{94.6 $\pm$ 1.0 }\\
EWC & 67.9 $\pm$ 2.0 & 78.1 $\pm$ 1.8 & 89.0 $\pm$ 1.6 & 94.4 $\pm$ 0.7 & 93.9 $\pm$ 0.6 \\
SGD & 65.9 $\pm$ 1.8 & 77.5 $\pm$ 1.5 & 88.6 $\pm$ 1.4 & \textbf{95.1 $\pm$ 0.5} & 94.1 $\pm$ 1.1 \\
SGD+Dropout & \textbf{81.1 $\pm$ 1.1}& \textbf{89.3 $\pm$ 2.4} & \textbf{92.1 $\pm$ 2.2} & 93.4 $\pm$ 1.8 & 92.8 $\pm$ 0.5 \\ \hline
\end{tabular}%
}
\caption{\textit{Rotated MNIST}: The validation accuracy of the model for each task, after being trained on all tasks in sequence.}\label{tab:rotated-mnist}
\end{table}

\section{Conclusion and Future Work}
In this paper, we studied the relationship between dropout and continual learning. We showed that the key to understanding this relationship is studying network stability. Furthermore, our analysis and experiments demonstrated that the dropout method could be viewed as an implicit gating mechanism, which yields a stable and plastic network. Our experiments showed that the consistent gating mechanism resulted from dropout can outperform various popular continual learning methods.

The effectiveness of the dropout method suggests that focusing directly on the stability of neural networks is an effective approach to tackle catastrophic forgetting. One interesting research direction is to modify the dropout method to gain more control over the gating mechanism, possibly by exploiting the structural similarity between sequential tasks and neural activation patterns, the same as proposed ideas in the transfer learning literature~\cite{Alinia2020ActiLabelAC}. Studying the effect of dropout on network behavior in different continual learning settings is also a promising direction. Our preliminary results show that dropout networks will remain robust even when trained on an increased number of sequential tasks.

\section*{Acknowledgement}
Authors Mirzadeh and Ghasemzdeh were supported in part, under grants CNS-1750679 and CNS-1932346 from the United States National Science Foundation. Any opinions, findings, conclusions, or recommendations expressed in this material are those of the authors and do not necessarily reflect the views of the funding organizations. The authors would like to thank the anonymous reviewers for their helpful comments.

{\small
\bibliographystyle{ieee_fullname}
\bibliography{refs}
}

\end{document}